\documentclass[letterpaper,twocolumn,fleqn]{article} 
\usepackage{ist}
\usepackage[utf8]{inputenc}
\usepackage[hyphens,spaces,obeyspaces]{url}
\usepackage[normalem]{ulem} %to strike the words
\usepackage{algorithm}
\usepackage{bm}
\usepackage{algorithmic}
\usepackage{graphicx}
\usepackage{amsmath}
\usepackage{multirow}
\usepackage{lipsum}
\usepackage{subcaption}
\title{Indoor Layout Estimation by 2D LiDAR and Camera Fusion}
\author{ Jieyu Li, Robert L. Stevenson; University of Notre Dame; Notre Dame, IN 46556.}
\date{} % date has an empty field.
\begin{document}
\maketitle
\begin{abstract} 
This paper presents an algorithm for indoor layout estimation and reconstruction through the fusion of a sequence of captured images and LiDAR data sets. In the proposed system, a movable platform collects both intensity images and 2D LiDAR information. Pose estimation and semantic segmentation is computed jointly by aligning the LiDAR points to line segments from the images. 
For indoor scenes with walls orthogonal to floor, the alignment problem is decoupled into top-down view projection and a 2D similarity transformation estimation and solved by the recursive random sample consensus (R-RANSAC) algorithm. Hypotheses can be generated, evaluated and optimized by integrating new scans as the platform moves throughout the environment. The proposed method avoids the need of extensive prior training or a cuboid layout assumption, which is more effective and practical compared to most previous indoor layout estimation methods. Multi-sensor fusion allows the capability of providing accurate depth estimation and high resolution visual information.

\end{abstract}
%\keywords{indoor layout estimation, 3D reconstruction, LiDAR and camera fusion, recursive random sample consensus}

\section{1 Introduction}

Spacial layout estimation of indoor scenes provides essential geometric information and constraints for various tasks, such as indoor 3D reconstruction, navigation, scene understanding and augmented reality. Most of the work in layout estimation ignores the clutter and decorations found in the environment, instead focusing more on the horizontal contours of walls. Such layout specifies segmentation and labels (e.g. walls, floor or wall-floor boundaries), and thus the surface orientation and depth estimates describing the spacial geometry can be estimated as well. 

The problem of layout reconstruction has been extensively studied using data from a single perspective image \cite{hoiem2005geometric, delage2007automatic, hedau2009recovering, lee2009geometric, liu2010single}, a single panorama image \cite{zou2018layoutnet, yang2016efficient}, multiple images \cite{tsai2011real, flint2011manhattan, cabral2014piecewise, zhang2019edge, phalak2019deepperimeter} and RGBD images \cite{liu2018floornet}. One group of these methods is based on supervised learning to obtain the semantic labels from multi-scale local features (e.g. color, texture, locations and lines), geometric features (e.g. edge orientation and vanishing points) \cite{hoiem2005geometric, delage2007automatic, liu2010single} and depth estimates from multiple images \cite{flint2011manhattan, cabral2014piecewise, phalak2019deepperimeter}. %?check reference 
However, real-time performance and generalization to different environments is difficult to achieve even with extensive training and inference at a superpixel level. 
%However, these methods are computationally intensive, making them difficult to apply in real time without specialized GPU hardware support \tsai

A second group of methods generates candidate layout hypotheses and ranks them with scoring functions. The generation rules of the candidates depends heavily on geometric assumptions, which limits the scope of these algorithms \cite{yang2016efficient}. For example, \cite{tsai2011real} considers the scene assembled by ground plane and at most three walls (i.e. left, end, and right walls) without occluding edges. \cite{hedau2009recovering} and \cite{zhang2019edge} generates layout hypotheses by coarsely sampling rays from vanishing points. %''Box" layout in \cite{hedau2009recovering} emphasizes ''Manhattan World" assumption as well, which states that all surfaces are aligned with three mutually orthogonal dominant directions. 
These assumptions prevent the algorithms from generalizing to more complex environments.  A more practical model is the ``indoor world'' model \cite{lee2009geometric} which describes multiple mutually orthogonal walls (``Manhattan World'' assumption) connected by convex, concave, and occluding corners. The ground-wall boundary is sufficient to give a complete 3D reconstruction of the ``indoor world'' model \cite{delage2007automatic}.

Most reconstruction methods also suffer from large distance errors especially for walls far away from the camera, and therefore fail for tasks in which geometry must be precise (e.g. floorplans or architectural blue-prints) \cite{liu2018floornet}. Even with RGBD cameras, the depth information is only accurate up to a very limited range (around 5m), and is generally noisy relative to the depth data available from LiDAR \cite{li20152d}.

Given these considerations, image and LiDAR fusion appears to be a promising solution. 2D LiDAR data can assist in defining the outline of rooms and help address corner connectivity properties. This can lead to more complex environments, beyond simple cuboid scenes. Common image and LiDAR fusion techniques \cite{li20152d, zhang2004extrinsic} %more reference
start with extrinsic calibration using manual placement of fiducial targets. Since the calibration error is evaluated by the Euclidean distance between the predicted and observed features, the effect along each direction in 3D space is not considered. As a result, the indoor reconstruction resulting from this method might fail to satisfy the geometric constraints (e.g. vertical walls and a single floor).

In this paper, a 2D LiDAR scanning parallel to the floor and a camera with fixed relative position are mounted on a moving platform. The semantic segmentation can be done by projecting the LiDAR points, as samples of the room contour, to identify the ground-wall boundary from line segments found in images. The alignment problem is decoupled into estimating the top-down homography and a 2D similarity transformation, which ensures the geometric constraints of vertical walls. Furthermore, due to the consistency of the relative position between the camera and the LiDAR, the best hypothesis should be able to project the LiDAR points to the ground-wall boundary in multiple scans. The method does not rely on any training from the image properties.

\section{2 Proposed Approach}
%\subsection{Overview of Approach}
%In the proposed system, a 2D LiDAR scanning the surface parallel to floor and a camera with fixed relative position are mounted on a robot moving on the floor.

\begin{figure*}[t]
    \centering
    \includegraphics[width=\textwidth]{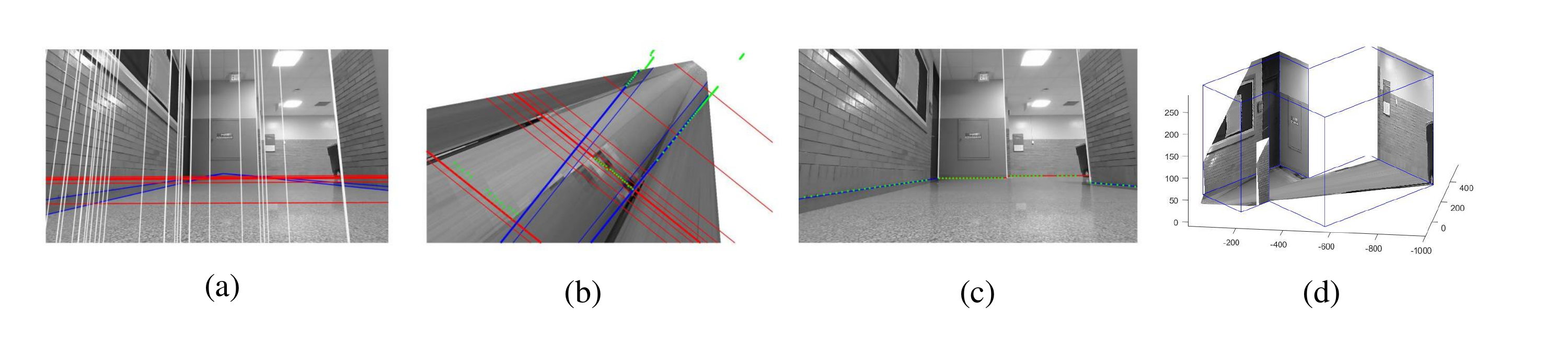}
    \caption{Illustration of the proposed system: (a) line and vanishing point detection (white lines corresponding to vertical lines, blue and red lines corresponding to two horizontal Manhattan directions relatively); (b) top-down projection of images and alignment between LiDAR points (green dots) and line segments; (c) semantic segmentation; (d) 3D reconstruction with image texture.}
    \label{overview}
\end{figure*}

The proposed method for constructing and evaluating the ground-wall boundary hypotheses is described in this section and illustrated by Fig. \ref{overview}. The proposed system consists of four modules: LiDAR data segmentation, line features detection, ground-wall boundary estimation by the alignment between LiDAR points and image lines, and 3D reconstruction.

%\begin{figure*}[h]
%\centering
%\begin{subfigure}[t]{0.4\textwidth}
%\centering
%\includegraphics[width=\textwidth]{figures/image.jpg}
%\caption{}
%\label{1(a)}
%\end{subfigure}
%\begin{subfigure}[t]{0.4\textwidth}
%\centering
%\includegraphics[width=\textwidth]{figures/align.jpg}
%\caption{}
%\label{1(b)}
%\end{subfigure}
%\begin{subfigure}[t]{0.4\textwidth}
%\includegraphics[trim=1cm 10cm 1cm 1cm, clip=true, width=\textwidth]{figure1.pdf}
%\centering
%\includegraphics[width=\textwidth]{figures/seg.jpg}
%\caption{}
%\label{1(c)}
%\end{subfigure}
%\begin{subfigure}[t]{0.4\textwidth}
%\centering
%\includegraphics[width=\textwidth]{figures/3d.jpg}
%\caption{}
%\label{1(d)}
%\end{subfigure}
LiDAR points are clustered into lines by Split-and-Merge method
(Sec. 2.1). Straight lines are extracted from the images and grouped into clusters for three mutually orthogonal vanishing points (Sec. 2.2). As shown in Fig. \ref{overview}(a), the ground-wall boundary candidates are composed by sets of the horizontal lines. 

The best hypothesis is identified if a transformation matrix aligning the LiDAR points and the boundaries can be found.
Sec. 2.3 describes the mapping process, which is decoupled into a top-down view projection and a 2D similarity transformation. Fig. \ref{overview}(b) shows an example of the ground-map with aligned LiDAR points in green. The ground-map, as the top-down view of the ground-wall boundary, is generated according to the vertical vanishing point.

Since the relative position of the LiDAR and the camera is fixed, this transformation matrix should be able to project most LiDAR points near the boundaries within a certain range (as inliers) in multiple scans. Thus the best hypothesis with the most inliers is identified and refined over the consensus set from multiple scans. This is solved by the recursive random sample consensus (R-RANSAC) algorithm, which is discussed in Sec. 2.4. %Multiple scans, captured by the moving platform, can be used to generate new hypotheses or update the existing model of the transformation.

As the last step, the semantic segmentation of images is done by choosing the horizontal lines near the projected LiDAR points as the ground-wall boundary and the vertical lines near the LiDAR corners as the wall edges (Fig. \ref{overview}(c)). With multiple scans, the location of the ground-wall boundary can be further refined by Extended Kalman Filter (EKF). The surface depth is given by the LiDAR data. The 3D reconstruction is shown in Fig. \ref{overview}(d)

\subsection{2.1 LiDAR points clustering}\label{lidarseg}
The LiDAR points are clustered into lines by the Split-and-Merge method \cite{nguyen2005comparison}. %The points are first grouped if the distance between two adjacent points is smaller than a threshold, and then a line segment is fitted in each cluster.
The points are first divided into clusters if the distance between two adjacent points is larger than a threshold, and then a line segment is fitted in each cluster. The line fitting can be simply done by setting the first and last points in the cluster to be the endpoints of the line, as in the Iterative-End-Point-Fit algorithm \cite{zhang2000line}. If the maximum deviation (perpendicular distance to the line) exceeds the allowable tolerance $\tau_{\rm l}$, the cluster is further divided at the point of maximum deviation. This testing process is repeated until the maximum deviation of all the line segments are smaller than $\tau_{\rm l}$. The LiDAR points clusters are further grouped into two sets corresponding to two horizontal directions according to the angles of the associated lines.

\subsection{2.2 Line and Vanishing Point Detection}
\label{line}
For line detection, the edges are first detected by the Canny edge detector. To group these edges into straight line segments, many methods have been developed based on Hough transform. However, this can result in unreasonable splitting or merging of lines caused by improper window size of the accumulator space. The algorithm in \cite{KovesiMATLABCode}, which avoids the need of careful setting of the window size, is implemented. The idea and process is very similar to the Split-and-Merge for LiDAR data processing (Sec. 2.1), but the initial grouping is done by connecting the sequential edge points up to the junction points. 

Given the detected lines, a set of vanishing points is estimated by \cite{tardif2009non}. The lines are assigned to a vanishing point or classified as outliers. This method is based on the J-linkage algorithm \cite{toldo2008robust}. It randomly chooses $M$ sets of two lines from $N$ lines and computes their intersection as vanishing point hypotheses. A preference matrix $\bm{Q}$, a $N\times M$ Boolean matrix, is defined with elements $\bm{Q}_{nm}$ describing if the $n^{\rm th}$ line is consistent to the $m^{\rm th}$ hypothesis (i.e. the distance between the vanishing point to the line is below a threshold). Based on the assumption that the lines corresponding to the same vanishing point tend to have similar preference sets (rows of $\bm{Q}$), the algorithm iteratively merges two clusters with minimal Jaccard distance until the preference sets of any two clusters are disjoint. In each iteration, The preference set of a cluster of lines is updated as the intersection of the preference sets of its members. Small clusters are classified as outliers, and three vanishing points are selected corresponding to the Manhattan directions. More details can be obtained in \cite{tardif2009non}.

\begin{figure*}
\centering
\begin{subfigure}[b]{0.3\textwidth}
    \includegraphics[width=\textwidth]{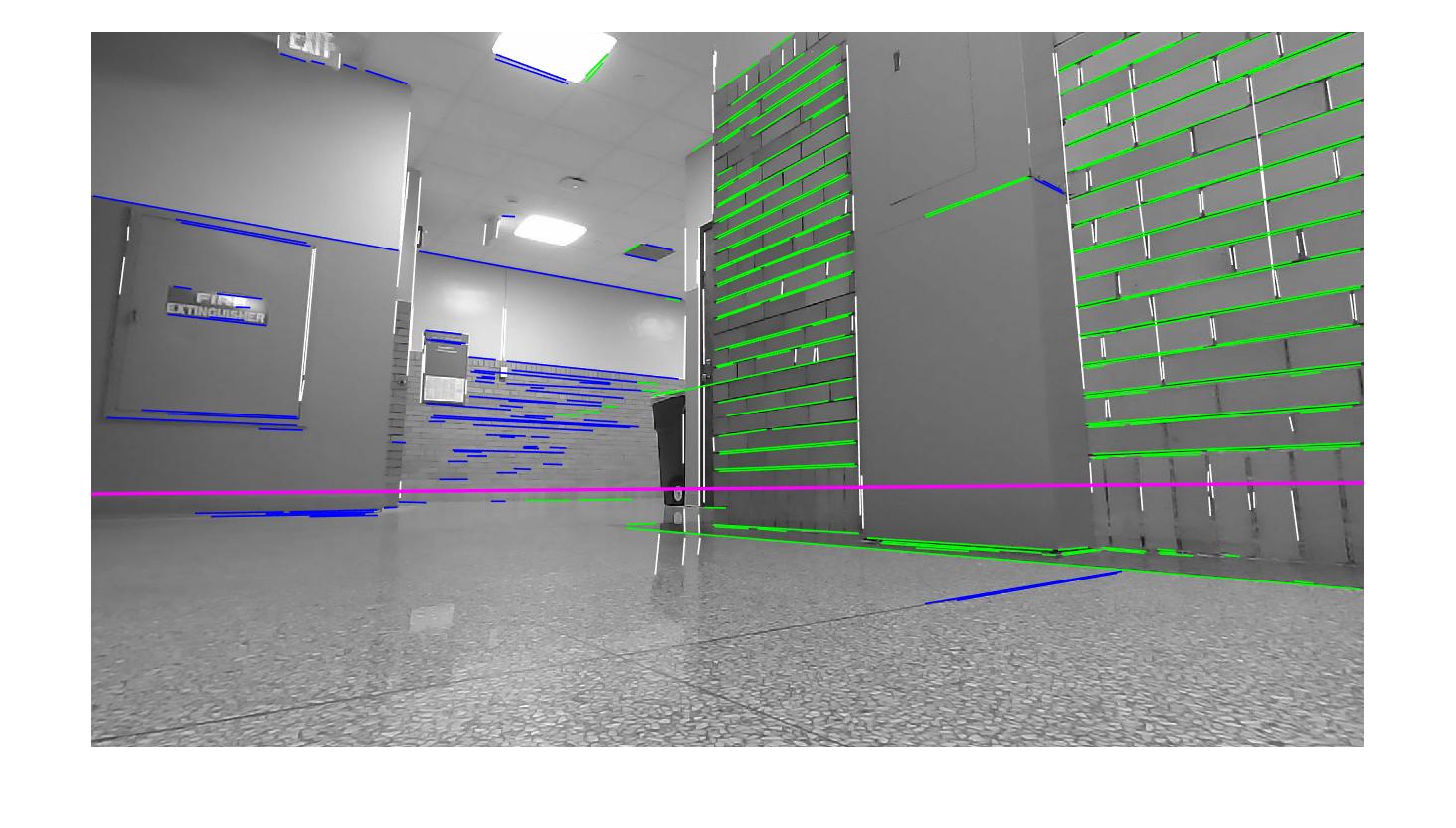}
%    \caption{}
 %   \label{linecluster}
\end{subfigure}
\begin{subfigure}[b]{0.3\textwidth}
    \includegraphics[width=\textwidth]{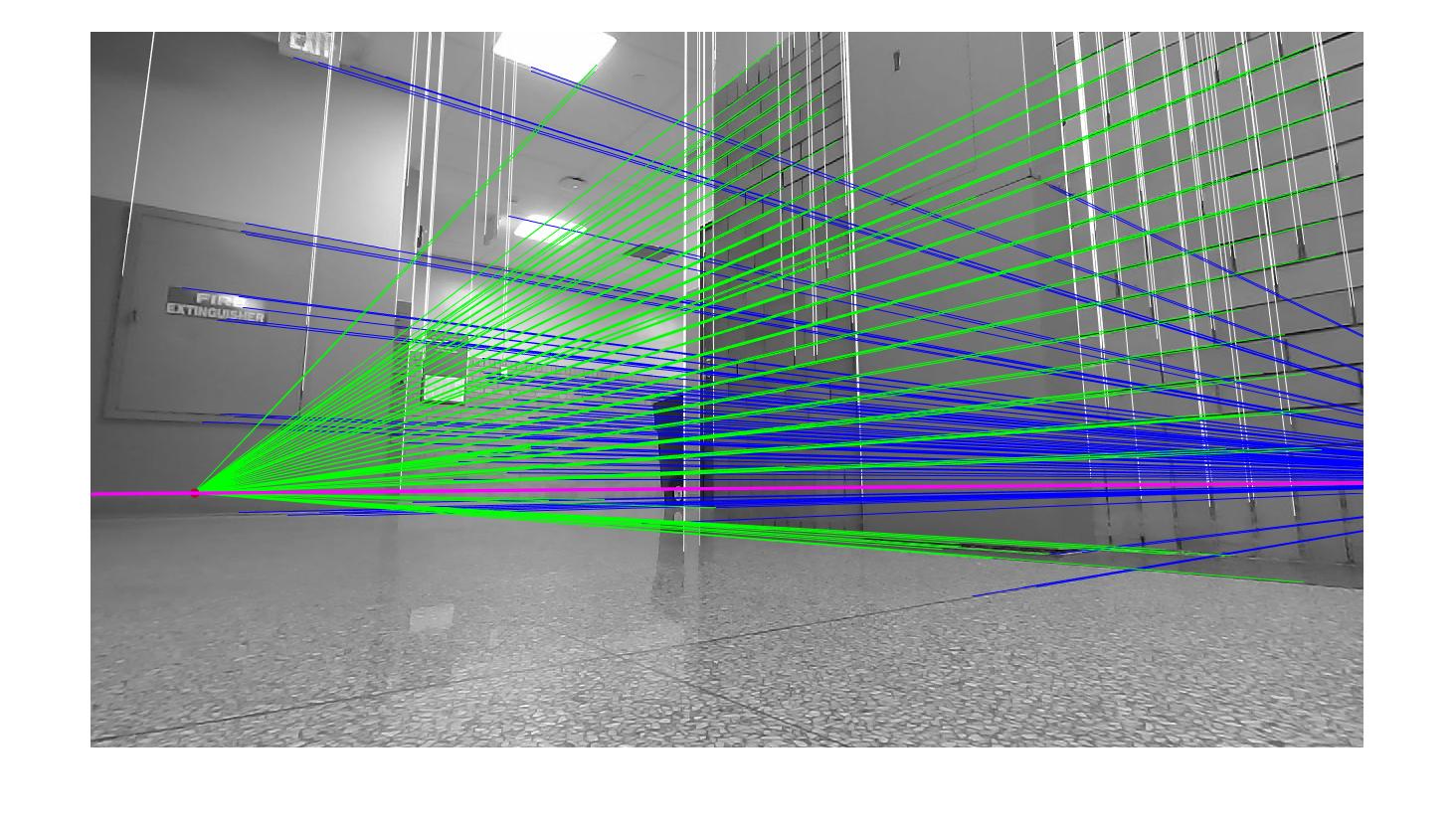}
    %\caption{}
    %\label{}
\end{subfigure}
\begin{subfigure}[b]{0.3\textwidth}
    \includegraphics[width=\textwidth]{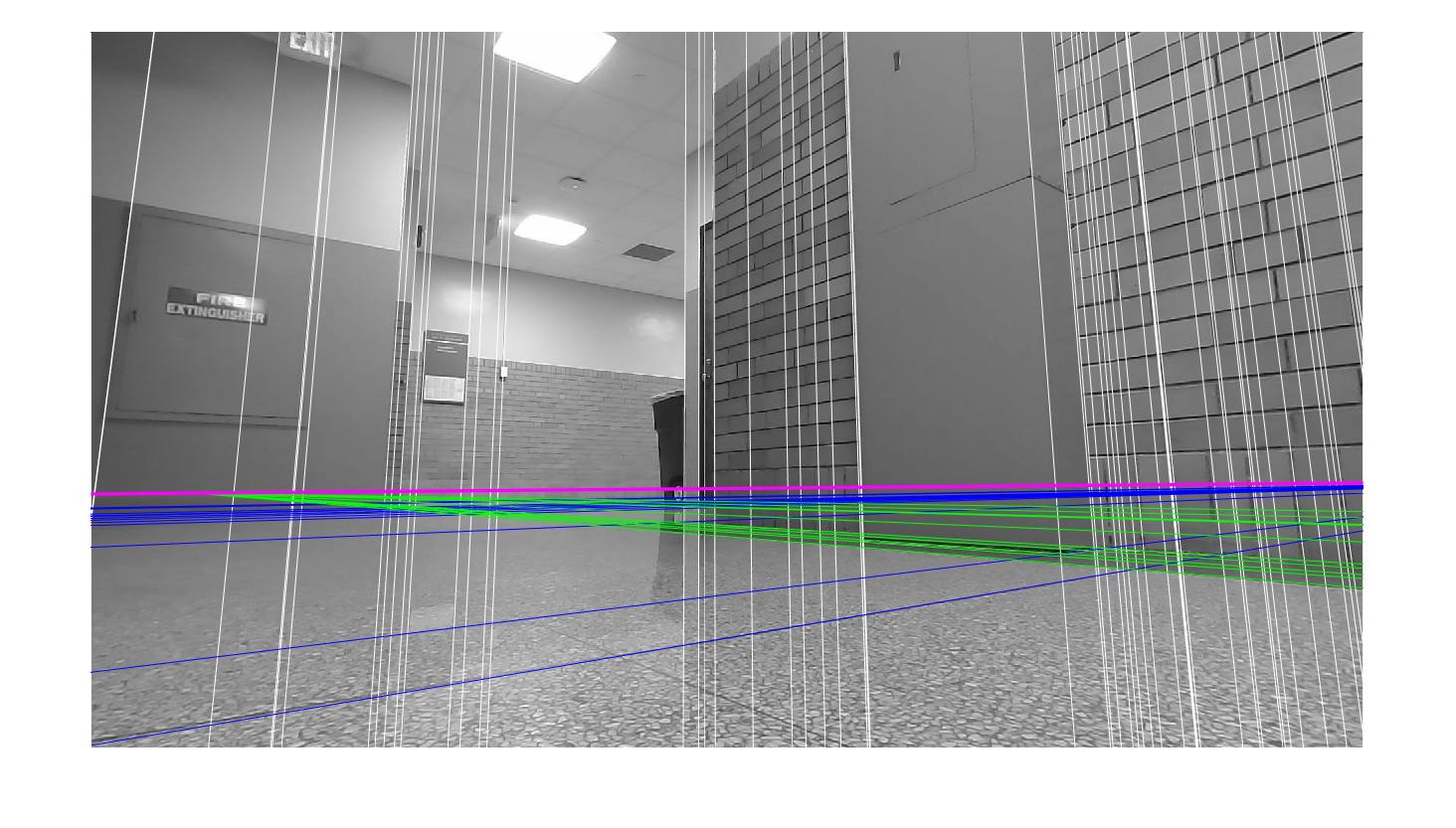}
    %\caption{}
    %\label{}
\end{subfigure}
\caption{Results of image data processing: (a) edge detection and clustering, (b) line and vanishing points detection, (c) ground-wall boundary candidates generation}
\label{imagedata}

\end{figure*}

%The lines are clustered by the J-linkage algorithm \cite{toldo2008robust}. The vanishing points corresponding to the Manhattan directions are estimated by \cite{tardif2009non}. Since the camera moves with fixed tilt and roll angle, the estimation of the vertical vanishing point can be refined with new scans captured. Orientation of a line segment is refined by setting the endpoints to be its centroid and corresponding vanishing point. 
An example of line and vanishing point detection is shown in Fig. \ref{imagedata}. Since the horizontal line is the vanishing line of the ground plane, the line segments with at least one endpoint being above the horizontal vanishing line are excluded from the ground-wall boundary candidates. The magenta line is the horizontal vanishing line; the blue and green lines correspond to two horizontal directions respectively; and the white lines are the vertical lines.

%\subsection{LiDAR data segmentation}
\subsection{2.3 LiDAR and Ground-wall Boundary Alignment}
\label{alignment}

The homography mapping can align the LiDAR points, as sampled from the room contour, to indicate the ground-wall boundary. The mapping is estimated by two steps: a homography matrix for top-down view of the image is calculated according to the vertical vanishing point; then a 2D similarity transformation is found to align the LiDAR points to the ground-map.

\subsubsection{2.3.1 Top-down view projection}

The method of top-down homograhpy estimation by vertical vanishing point is described in this section. More detailed properties of vanishing points can be found in \cite{caprile1990using}. 

A vanishing point is a point on the image plane where the 2D perspective projections of mutually parallel lines in 3D space appear to converge. If $\bm{s}$ is a straight line in 3D space with direction $\bm{n}_s=[n_{\rm x},n_{\rm y},n_{\rm z}]^{\rm T}$ through a point $\bm{p}_a$, points on the line $\bm{s}$ can be given by
\begin{equation}
\bm{p}_s=\bm{p}_a+t\bm{n}_s=\begin{bmatrix}x_a \\ y_a\\ z_a\end{bmatrix}+t\begin{bmatrix}n_{\rm x} \\ n_{\rm y}\\ n_{\rm z}\end{bmatrix}.
\end{equation}
Following the perspective projection model of camera, the vanishing point of $\bm{s}$ is determined by
\begin{equation}
\bm{v}_s=\lim\limits_{t \to \infty}  \begin{bmatrix} f \frac{x_a+tn_{\rm x}}{z_a+tn_{\rm z}} \\  f \frac{y_a+tn_{\rm y}}{z_a+tn_{\rm z}}\\ f\end{bmatrix}=\begin{bmatrix}f\frac{n_{\rm x}}{n_{\rm z}} \\ f\frac{n_{\rm y}}{n_{\rm z}}\\ f\end{bmatrix},
\end{equation}
where $f$ is the focal length. This gives the fact that if the vanishing point $\bm{v}_s$ of a straight line is known, the direction $\bm{n}_s$ of the line can be calculated as the unit vector associated with $\bm{v}_s$.

A coordinate system can be defined by a set of three mutually orthogonal unit vectors as the axes, following the right-hand rule. Let the associated vanishing points be $\bm{v}_Q=[x_Q, y_Q, f]^{\rm T}$, $\bm{v}_R=[x_R, y_R, f]^{\rm T}$ and $\bm{v}_S=[x_S, y_S, f]^{\rm T}$, the orthogonality gives \begin{equation}
 %   \left\{  
    \begin{array}{l}  
    x_Qx_R+y_Qy_R+f^2=0\\  
    x_Qx_S+y_Qy_S+f^2=0.\\  
    x_Rx_S+y_Ry_S+f^2=0
    \end{array}
\label{vanishingpoint}
\end{equation}
Thus by exploiting the orthogonality, the coordinates of the vanishing points and thus the axes of the coordinate system can be determined if at least three variables in Equation \ref{vanishingpoint} are known.

For top-down view projection, an arbitrary coordinate system is defined with the ${\rm x-y}$ plane as the ground plane and the ${\rm z}$ axis parallel to the vertical walls and pointing to the ground.
%With the camera moving with fixed tilt and roll angle, the vertical vanishing point, $v_{\rm g}=[x_{\rm g},y_{\rm g},f]^{\rm T}$, can be estimated and refined with multiple images. 
The vertical direction, as the unit vector associated to the vertical vanishing point, can be estimated by images. Since the camera is tilted downward, $y_{\rm v}<0$ gives an associated direction pointing to ceiling.
%The vertical vanishing point, $v_{\rm g}=[x_{\rm g},y_{\rm g},f]^{\rm T}$, can be estimated by images. 
By setting $\bm{v}_Q
=[x_{\rm v},y_{\rm v},f]^{\rm T}$ and $y_R=0$ in Equation \ref{vanishingpoint}, the axes of the arbitrary coordinate system in the camera frame can be defined as 
%With the vanishing point found, the axes are defined as associated unit vector $n_{\rm g}=[x_{\rm g}/h_z,y_{\rm g}/h_{\rm g},f/h_{\rm g}]^{\rm T}$, where $h_{\rm g}=\sqrt{x_{\rm g}^2+y_{\rm g}^2+f^2}$.
\begin{equation}
\begin{aligned}
    \begin{array}{l}
    \bm{e}_{\rm x}=\frac{1}{h_x}[-\frac{f^2}{x_{\rm v}},0,f]^{\rm T} \\
    \bm{e}_{\rm y}=\pm\frac{1}{h_y}[x_{\rm v},-\frac{x_{\rm v}^2+f^2}{y_{\rm v}},f]^{\rm T}, \\
    \bm{e}_{\rm z}=-\frac{1}{h_{\rm v}}[x_{\rm v},y_{\rm v},f]^{\rm T}\\
    \end{array}  
%    \\ & \text{subject to} \quad n_x \times n_y = -n_z
\end{aligned}
\end{equation}
where $h_{\rm x}$, $h_{\rm y}$ and $h_{\rm v}$ are the normalization parameters. Note that, the sign of $\bm{e}_{\rm y}$ is chosen so that coordinate system follows the right-hand rule, i.e. $\bm{e}_{\rm x} \times \bm{e}_{\rm y} = \bm{e}_{\rm z}$.

The rotation matrix for the top-down view can be calculated according to ${\rm \bm{I}}=\bm{R}[\bm{e}_{\rm x},\bm{e}_{\rm y},\bm{e}_{\rm z}]$, where ${\rm \bm{I}}$ is the identity matrix. 
So the rotation matrix is
\begin{equation}
%\begin{aligned}
%    R= R_xR_{\rm w} & =\begin{bmatrix}1 &0 &0\\0 &-1 &0\\0 &0 &-1 \end{bmatrix}[\bm{n}_x,\bm{n}_y,\bm{n}_z]^{\rm T}\\& =[\bm{n}_x,-\bm{n}_y,-\bm{n}_z]^{\rm T}
    \bm{R} =[\bm{e}_x,\bm{e}_y,\bm{e}_z]^{\rm T}.
%\end{aligned}
\end{equation}

A point $\bm{P}^{\rm c}$ in camera coordinate system is projected on the image plane with pixel coordinate $\bm{p}=[u, v, 1]^{\rm T}$ following the perspective projection model $\bm{p}\propto \bm{K}\bm{P}^{\rm c}$, where $\bm{K}$ is the intrinsic matrix of the camera. The coordinate of the rotated point in the top-down view, $\bm{p}^{\rm g}=[x^{\rm g}, y^{\rm g}, 1]^{\rm T}$, can be calculated by
\begin{equation}
 %\begin{bmatrix}x^{\rm g}\\y^{\rm g}\\1 \end{bmatrix} \propto KR\begin{bmatrix}X^{\rm c}\\Y^{\rm c}\\Z^{\rm c} \end{bmatrix}
 %\propto KRK^{-1}\bm{p}.
 \bm{p}^{\rm g}\propto \bm{KR}\bm{P}^{\rm c} \propto \bm{KRK}^{-1}\bm{p}.
\end{equation}
Thus the homography matrix of top-down projection is $\bm{H}=\bm{KRK}^{-1}$.

%In order to get the top-down perspective, the camera needs to be rotated to let the optical axis pointing to the floor.
%\begin{equation}
%    I=R[n_x,n_y,-n_z],
%\end{equation}
%where $R$ is the rotation matrix, and $I$ is the 3-by-3 identity matrix, 

\subsubsection{2.3.2 Similarity transformation}

The LiDAR points and the ground-map can be aligned by a 2D similarity transformation. Four parameters are needed to be estimated, including rotation angle $\phi$, translation $(t_{\rm x},t_{\rm y})$ and a isotropic scaling $\delta$ (invariant scaling with respect to direction). With the minimum subsets, any two associated pairs of points from ground-map $\bm{p}^{\rm g}_j=[x^{\rm g}_j, y^{\rm g}_j]^{\rm T}$ and LiDAR data $\bm{p}^{\rm l}_j=[x^{\rm l}_j, y^{\rm l}_j]^{\rm T}$ ($j=1,2$), the unique solution of these parameters can be calculated by
\begin{equation}
\begin{array}{l}
\delta =\frac{\sqrt{\Delta {y^{\rm l}}^2+\Delta {x^{\rm l}}^2}}{\sqrt{\Delta {y^{\rm g}}^2+\Delta {x^{\rm g}}^2}}\\
\phi= {\rm atan2}(\Delta y^{\rm l}, \Delta x^{\rm l})-{\rm atan2}(\Delta y^{\rm g}, \Delta x^{\rm g}),\\
t_{\rm x} = \bar{x}^{\rm l}-\bar{x}^{\rm g}\delta\cos\phi+\bar{y}^{\rm g}\delta\sin\phi\\ %\frac{1}{2}[x_1^{\rm l}+x_2^{\rm l}-k\cos{\rho}(x_1^i+x_2^i)-k\sin{\rho}(y_l^i+y_2^i)]\\
t_{\rm y} =\bar{y}^{\rm l}-\bar{x}^{\rm g}\delta\sin{\phi}-\bar{y}^{\rm g}\delta\cos{\phi}
\end{array}  
\label{similaritytrans}
\end{equation}
where $\Delta$ indicates the difference (i.e. $\Delta x^{\rm l}=x^{\rm l}_1-x^{\rm l}_2$), %, $\Delta y_l=y_{l,1}-y_{l,2}$, $\Delta x_i=x_{i,1}-x_{i,2}$ and $\Delta y_i=y_{i,1}-y_{i,2}$;
and  $\bar{\bm{x}}$  indicates the average value of $\bm{x}$.% (i.e. $\bar{x}^{\rm l}=\frac{1}{2}(x^{\rm l}_1+x^{\rm l}_2)$).
%$\bar{x_l}=\frac{1}{2}(x_{l,1}+x_{l,2})$, $\bar{y_l}=\frac{1}{2}(y_{l,1}+y_{l,2})$, $\bar{x_i}=\frac{1}{2}(x_{i,1}+x_{i,2})$, $\bar{y_i}=\frac{1}{2}(y_{i,1}+y_{i,2})$

Overdetermined sets of equations defined by more associated points can provide a more robust solution by minimizing a suitable cost function (e.g. algebraic distance or geometric distance). 

\subsection{2.4 Ground-wall Boundary Estimation with Multiple Scans}
\label{ransac}

The proposed method needs to identify the ground-wall boundary from many other horizontal line segments detected in images. The random sample consensus (RANSAC) algorithm is a robust approach for model estimation and data association with a large number of spurious measurements. 

The RANSAC algorithm first forms many hypotheses using minimum subsets of measurements. As shown by Equation \ref{similaritytrans}, the minimum subsets for estimating the 2D similarity transformation can be any two pairs of corners, given by the intersections of randomly selected lines associated to two directions respectively from the ground-map and the LiDAR data.

The best hypothesis with the most inliers is identified and refined. Given the $k^{\rm th}$ hypothesis $\hat{\bm{\theta}}_k$, inliers of each scan are the LiDAR points in sets that the average distance to the associated ground-map line is below a threshold $\tau_{\rm d}$. 
%To evaluate the hypotheses, the first step is to associate the projected LiDAR points to the nearest image lines.
The ground-map line $\bm{s}^{\rm g}_i$, associated to the $i^{\rm th}$ LiDAR points set $S^{\rm l}_i$, is assigned by minimizing the sum of distance between lines and all the points in $S^{\rm l}_i$.
%the associated line $s^{\rm g}_i$ of the LiDAR point  is found by the following equation.
\begin{equation}
%\begin{aligned}
\bm{s}^{\rm g}_i= \arg\min_{\bm{s}}\sum_j^{\bm{p}^{\rm l}_j \in S^{\rm l}_i}d(\bm{s},t(\bm{p}^{\rm l}_j,\hat{\bm{\theta}}_k)),
%\text{subject to }  \forall	
%\end{aligned}
\end{equation}
where $d(\cdot)$ calculates the perpendicular distance from a point to a line, $t(\cdot)$ indicates the top-down projection and similarity transformation. 

Furthermore, the ``Manhattan World'' constraint is considered in the process of data association. With the LiDAR data and image lines which are grouped into two sets corresponding to the two horizontal directions, there are two possible combination for the data association, such as matching the first LiDAR set to one set of the image lines and the second LiDAR set to the other, and vice versa. The combination that gives most inliers is chosen.

%\begin{equation}
%\begin{aligned}
%    C_k := \{p^{\rm l}_i|d(s^{\rm g}_i,t(p^{\rm l}_i,\hat{\bm{\theta}}_k))<\tau_d, i=1,2,...L\},
%\end{aligned}
%\label{conset}
%\end{equation}
%where $L$ is the number of LiDAR points.
%and $M$ are the number of scans and the LiDAR points in each scan respectively. 

With multiple scans, the similarity transformation hypotheses as well as the top-down homography estimation can be refined when data of a new scan is received. The recursive-RANSAC algorithm extends the basic RANSAC algorithm by storing multiple hypotheses and recursively updating them based on sequential measurements \cite{niedfeldt2014multiple}. R-RANSAC tests each new scan if enough inliers can be found according to any existing hypotheses. If so, those hypotheses are updated on their consensus sets; else instead of discarding the measurements, they are used to seed a new hypothesis.

The consensus set for the $k$-th hypothesis includes the scans that the inliers percentage is above a certain threshold. The top-down projection matrix can be smoothed according to the average vertical vanishing point weighted by the number of vertical lines in each scan. And the parameters for 2D similarity transform are refined by minimizing the distance between the inlier points and their associated ground-map lines over the consensus set. The pseudocode of the R-RANSAC algorithm can be found below.

With the transformation matrix, the location of the ground-wall boundary can be further refined by Extended Kalman Filter (EKF) with lines near the projected LiDAR points in each scan as the observation. If the robot pose is known, similar method can be found in \cite{tsai2012dynamic}, otherwise, the robot pose can be estimated simultaneously by SLAM \cite{choi2008line}.

\begin{algorithm}[!t]
\caption{R-RANSAC}
\begin{algorithmic}
\STATE Input: Observations, number of iterations $N$, error threshold $\tau$.\\
Initialization: nubmer of hypothese $k=0$.
\FOR{each new scan}
    \STATE Get the number of inliers according to each hypothesis $m(i)$ ($i=1,2,...,k$).
    \IF{$\forall m<\tau$}
        \STATE $k=k+1$ .
        \FOR{iteration $n<N$}
            \STATE Randomly select a minimum subset.\\
            Generate a hypothesis by Eq. \ref{similaritytrans}. \\
            Find the inliers.
            \IF{number of inliers is larger than previous iterations}
                \STATE Update the hypothesis $\hat{\bm{\theta}}_{k}$.
            \ENDIF
        \ENDFOR
        \STATE Optimize $\hat{\bm{\theta}}_k$ on the consensus set.
    \ELSE
    \STATE Optimize $\hat{\bm{\theta}}_{i}$ with $m(i)>\tau$ on its consensus set.
    \ENDIF
    \STATE Update the consensus set for each hypothesis.
\ENDFOR
\STATE Identify the best hypothesis.
\end{algorithmic}
\end{algorithm}

%update of the vanishing points and line segments to meet lidar points

\section{3 Experiments and Results}

\begin{figure*}[!thb]
\centering
\includegraphics[trim={0 10 0 0},clip, width=0.9\textwidth]{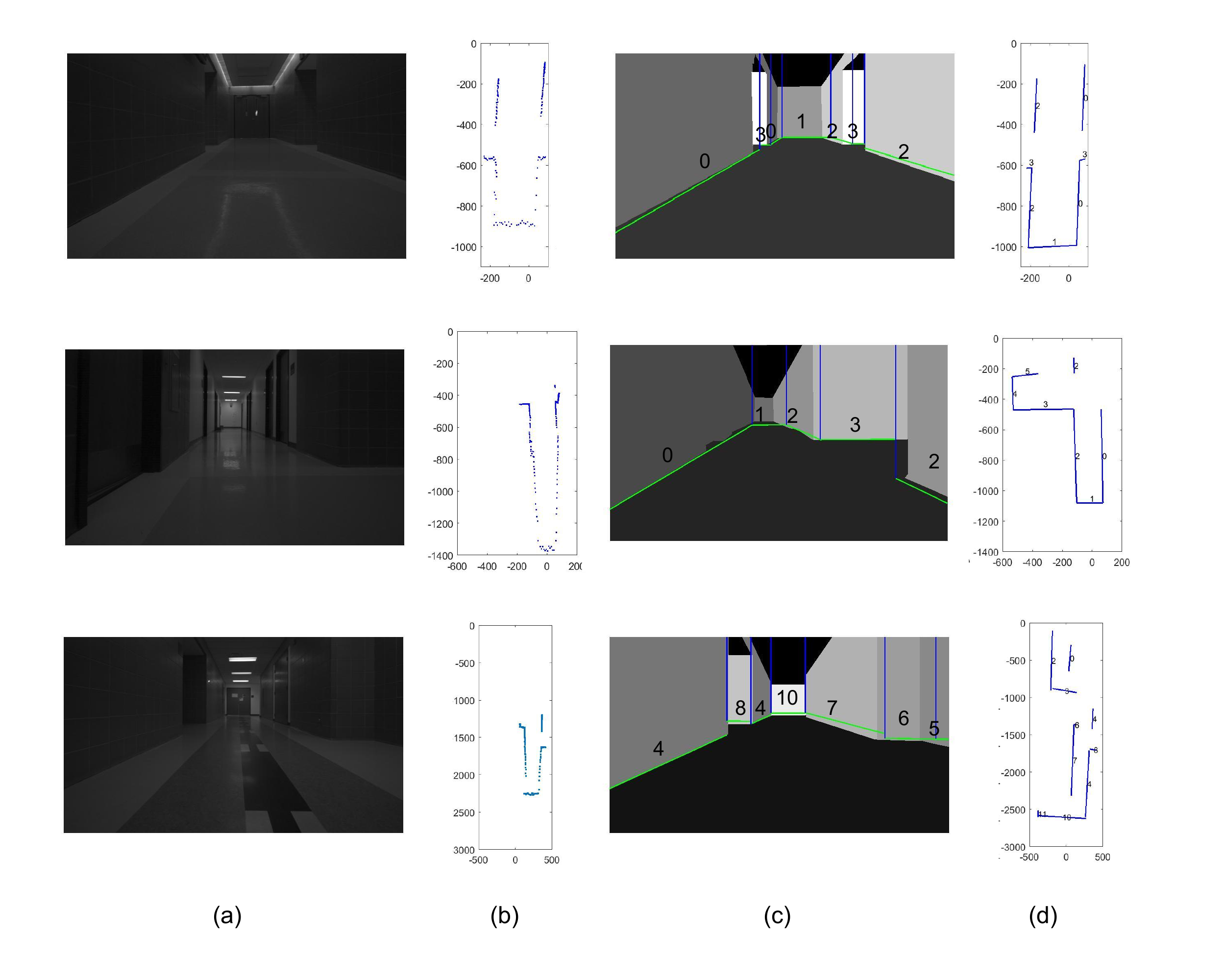}
\caption{Examples of data and test result of ``Michigan Indoor Corridor Dataset'' (The first, second and third rows are from ``Dataset +'', ``Dataset T1'' and ``Dataset L'' respectively). (a) is the image data; (b) is the LiDAR data sampled from the groundtruth and corrupted by noise (in cm); (c) shows the comparison between the groundtruth and the estimation result (indexs are the wall labels); and (d) is the ground-map (in cm) estimated by EKF. }
\label{umich}
\end{figure*}

The performance of the proposed approach is tested 
using two sets of data, the ``Michigan Indoor Corridor Dataset'' \cite{tsai2012dynamic}\footnotemark{} and a set of images and LiDAR data captured at the University of Notre Dame. The ``Michigan Indoor Corridor Dataset'' consists of sets of image sequences in various indoor environments collected by a calibrated camera mounted with zero tilt and roll angles and known fixed height. These specific camera setup constraints is not necessary for the proposed method. The ground truth images consist the labels of planes, i.e. the walls (numbered with an incremental counter), ground and ceiling plane. Although the range data was used for pose estimation in \cite{tsai2012dynamic}, it is not provided in this dataset. So the LiDAR data is generated from the ground truth by sampling the room contour per degree and distorted with 
Gaussian noise. 
Some other work tested on this dataset can be found in \cite{furlan2013free}. Fig. \ref{umich} shows some examples of results.

The quantitative results are reported in Table \ref{result}. The classification accuracy is the percentage of correctly labeled pixels according to the estimated indoor layout, without considering the ceiling pixels. 

\begin{table*}%{0.8\textwidth}
\begin{center}\begin{tabular}{ |c|c|c|c|c| }
\hline
Dataset & Dataset L & Dataset + & Dataset T1  & Overall\\
\hline
Proposed method &95.64\%  & 97.24\% & 98.09\% & 96.59\%\\
%Our method &95.64\%  & 96.18\% & 93.61\%\\
\hline
\cite{tsai2012dynamic} & 91.00\% & 94.23\% & 92.71\% &92.12\%\\
\hline
number of scans & 90 & 30 & 41 &167\\
\hline
\end{tabular}
\end{center}
\caption{Classification accuracy on the Michigan Indoor Corridor Dataset. Our results are compared to the results obtained with \cite{tsai2012dynamic}.}
\label{result}
\end{table*}
\footnotetext{The Michigan Indoor Corridor Dataset is available on \url{https://deepblue.lib.umich.edu/data/concern/data\_sets/3t945q88k}, but of which the Dataset T2 is corrupted.}

To the best of our knowledge, the similar camera and 2D LiDAR setup, and thus directly comparable work, has not been found. However, the comparison between \cite{tsai2012dynamic}, which uses LiDAR data only for pose estimation, shows that the proposed method can achieve better performance by combining LiDAR data and image sequences for indoor layout estimation. Furthermore, the proposed method needs less scans to learn a certain area of indoor environments, compared to the methods based on tracking image features, since large overlapping area of two subsequent scans is not necessary.

Additional tests were done at the University of Notre Dame. The images were taken in a hallway and around a corner of the building respectively. Each set contains 15 to 20 scans with the overall motion about 5 meters. The LiDAR data was captured by Slamtec's laser range scanner ``RPLIDAR A3''.
\begin{figure*}
\centering
\includegraphics[trim={0 120 0 0},clip, width=\textwidth]{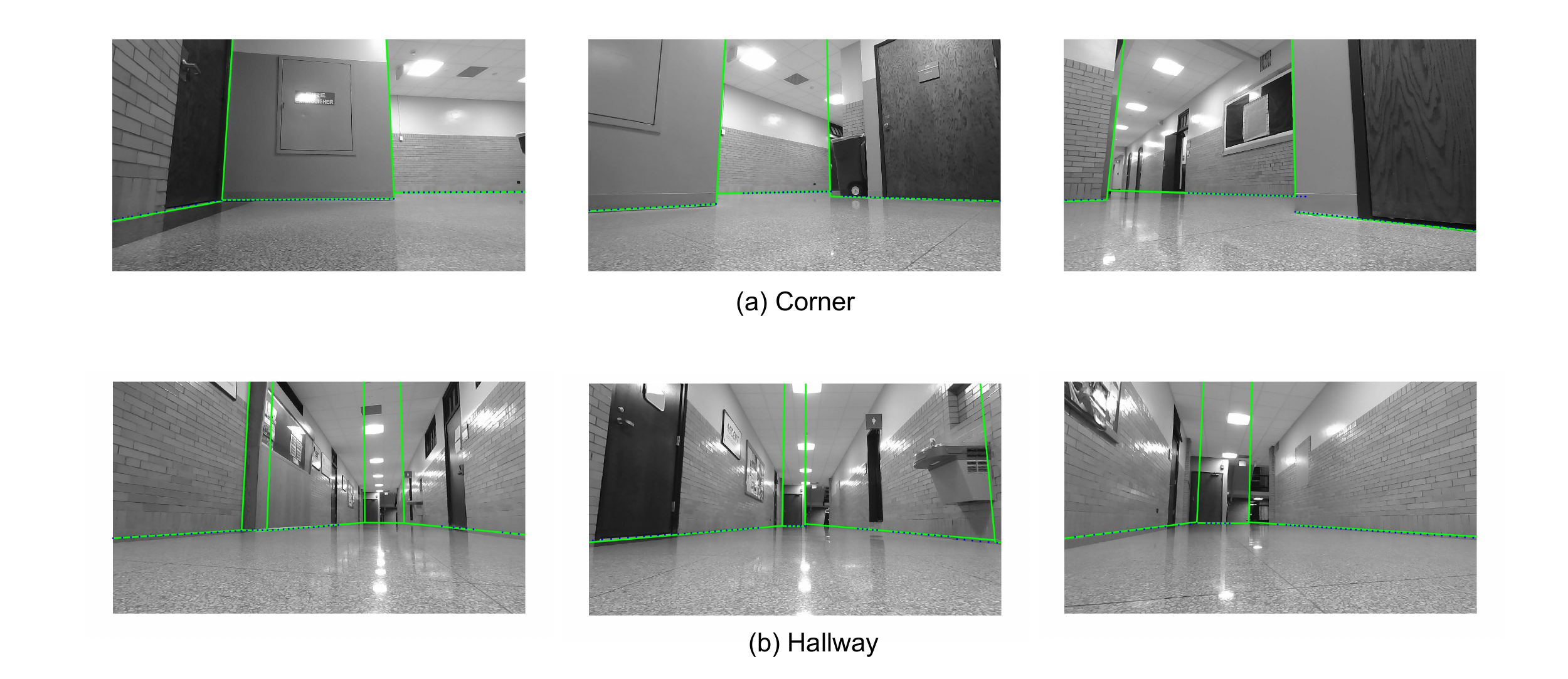}
\caption{Semantic segmentation results of different environment}
\label{nd}
\end{figure*}

As shown in Fig. \ref{nd}(b), the end wall of the hallway might be out of range of the LiDAR. In this case,
the reconstructed location of the end wall, defined by  
the furthest line segment below the horizontal
vanishing line, is not accurate. However, since the platform is moving towards the end wall, the location of the end wall is updated and better precision can be get by EKF.
%Furthermore, as in offices, the ground-wall boundary is partially or entirely occluded by clutters, which may lead to large layout reconstruction error. As a result, method to identify and exclude the clutters from LiDAR points is needed in the future work.

\section{4 Conclusion}
An algorithm for indoor layout estimation based on the  fusion of a sequence of images and 2D LiDAR data is presented in this paper.
The ground-wall boundaries in images are identified by 
projecting the LiDAR points, as sampled from the contour of the rooms. The alignment problem is decoupled into top-down view projection and a 2D similarity transformation estimation. Multiple scans are used to refine both the alignment estimation by the R-RANSAC algorithm and the location of walls in ground map by EKF.

The average classification accuracy, tested on the ``Michigan Indoor Corridor Dataset'', is $96.59\%$, higher than the accuracy of most State-of-the-art approaches\cite{furlan2013free}. The LiDAR and camera fusion provides both accurate depth estimation and high resolution visual information. The proposed method avoids the need of extensive prior training or a cuboid layout assumption. Furthermore, with a wide field of view and the ability to guide visual features tracking, LiDAR, combined with camera, can be a promising solution for simple and accurate 3D reconstruction of large scenes.


\small
\begin{thebibliography}{9}
\bibitem{cabral2014piecewise}Ricardo Cabral and Yasutaka Furukawa, Piecewise Planar Andcompact Floorplan Reconstruction from Images, In 2014 IEEE Conference on Computer Vision and Pattern Recognition, pg. 628–635. IEEE, (2014).
\bibitem{caprile1990using}Bruno Caprile and Vincent Torre, Using Vanishing Points for Camera Calibration, International Journal of Computer Vision, 4(2):127–139, (1990).
\bibitem{choi2008line}Young-Ho Choi, Tae-Kyeong Lee and Se-Young Oh, A Line Feature Based SLAM with Low Grade Range Sensors Using Geometric Constraints and Active Exploration for Mobile Robot, Autonomous Robots, 24(1):13–27, (2008).
\bibitem{delage2007automatic}Erick Delage, Honglak Lee and Andrew Y Ng,  Automatic Single-image 3D Reconstructions of Indoor Manhattan World Scenes,  In Robotics Research, pg. 305–321, Springer, (2007).
\bibitem{flint2011manhattan}Alex Flint, David Murray and Ian Reid, Manhattan Scene Understanding Using Monocular, Stereo, and 3D Features, In 2011 Inter-national Conference on Computer Vision, pg. 2228–2235, IEEE, (2011).
\bibitem{furlan2013free}Axel Furlan, Stephen D Miller, Domenico G. Sorrenti, Fei-Fei Li and Silvio Savarese, Free your Camera: 3D Indoor Scene Understanding from Arbitrary Camera Motion, In BMVC, (2013).
\bibitem{hedau2009recovering}Varsha Hedau, Hoiem Derek and David Forsyth, Recovering the Spatial Layout of Cluttered Rooms, In 2009 IEEE 12th International Conference on Computer Vision,pg. 1849-1856, IEEE, (2009).
\bibitem{hoiem2005geometric} Derek Hoiem, Alexei A Efros and Martial Hebert, Geometric Context from a Single Image, In 10th IEEE International Conference on Computer Vision (ICCV'05), Volume 1, pg. 654-661, IEEE, (2005).
\bibitem{KovesiMATLABCode}P. D. Kovesi, MATLAB and Octave Functions for Computer Vision and Image Processing, Available from: $<$http://www.peterkovesi.com/matlabfns/$>$
\bibitem{lee2009geometric}David C Lee, Martial Hebert and Takeo Kanade, Geometric Reasoning for Single Image Structure Recovery, In 2009 IEEE Conference on Computer Vision and Pattern Recognition, pg. 2136-2143, IEEE, (2009).
\bibitem{li20152d}Juan Li, Xiang He, and Jia Li. 2D LiDAR and Camera Fusion in 3D Modeling of Indoor Environment, In 2015 National Aerospace and Electronics Conference (NAECON), pg. 379–383. IEEE, (2015).
\bibitem{liu2010single}Beyang Liu, Stephen Gould, and Daphne Koller, Single Image Depth Estimation from Predicted Semantic Labels, In 2010 IEEE Computer Society Conference on Computer Vision and Pattern Recognition, pg. 1253–1260, IEEE, (2010).
\bibitem{liu2018floornet}Chen Liu, Jiaye Wu, and Yasutaka Furukawa, Floornet: A unified Framework for Floorplan Reconstruction from 3d Scans, In Proceedings of the European Conference on Computer Vision(ECCV), pg. 201–217, (2018).
\bibitem{nguyen2005comparison}Viet Nguyen, Agostino Martinelli, Nicola Tomatis and Roland Siegwart, A Comparison of Line Extraction Algorithms Using 2D Laser Rangefinder for Indoor Mobile Robotics, In 2005 IEEE/RSJInternational  Conference  on  Intelligent  Robots  and  Systems, pg. 1929–1934, IEEE, (2005).
\bibitem{niedfeldt2014multiple}Peter C Niedfeldt and Randal W Beard, Multiple Target Rracking Using Recursive RANSAC, In 2014 American Control Conference, pg.3393–3398, IEEE, (2014).
\bibitem{phalak2019deepperimeter}Ameya Phalak, Zhao  Chen, Darvin Yi, Khushi Gupta, Vijay Badrinarayanan and Andrew Rabinovich, Deepperimeter: Indoor Boundary  Estimation from Posed Monocular Sequences, arXiv preprint arXiv: 1904.11595, (2019).
\bibitem{tardif2009non}Jean-Philippe Tardif, Non-iterative Approach for Fast and Accurate Vanishing Point Detection, In 2009 IEEE 12th International Conference on Computer Vision, pg. 1250–1257, IEEE, (2009).
\bibitem{toldo2008robust}Roberto Toldo and Andrea Fusiello, Robust Multiple Structures Estimation with J-linkage, In European Conference on Computer Vision, pg. 537–547, Springer, (2008).
\bibitem{tsai2012dynamic}Grace Tsai and Benjamin Kuipers, Dynamic Visual Understanding of the Local Environment for an Indoor Navigating Robot, In 2012 IEEE/RSJ International Conference on Intelligent Robots and Systems, pg. 4695–4701, IEEE, (2012).
\bibitem{tsai2011real}Grace Tsai, Changhai Xu, Jingen Liu and Benjamin Kuipers, Real-time Indoor Scene Understanding Using Bayesian Filtering with Motion Cues, In ICCV, pg.121–128, (2011).
\bibitem{yang2016efficient}Hao Yang and Hui Zhang, Efficient 3D Room Shape Recovery from a Single Panorama, In Proceedings of the IEEE Conference on Computer Vision and Pattern Recognition, pg. 5422–5430, (2016).
\bibitem{zhang2000line}Li Zhang and Bijoy K Ghosh, Line Segment Based Map Building and Localization Using 2D Laser Rangefinder, In Proceedings 2000 ICRA. Millennium Conference. IEEE International Conference on Robotics and Automation, Symposia Proceedings (Cat. No.00CH37065), volume 3, pg. 2538–2543, IEEE, (2000).
\bibitem{zhang2004extrinsic}Qilong Zhang and Robert Pless, Extrinsic Calibration of a Camera  and Laser Range Finder (Improves Camera Calibration), In 2004 IEEE/RSJ International Conference on Intelligent Robots and  Systems (IROS) (IEEE Cat. No. 04CH37566), volume 3, pg. 2301–2306, IEEE.
\bibitem{zhang2019edge}Weidong  Zhang, Wei Zhang and Jason Gu, Edge-semantic Learning Strategy for Layout Estimation in Indoor Environment, IEEE transactions on cybernetics, (2019).
\bibitem{zou2018layoutnet}Chuhang Zou, Alex Colburn, Qi Shan and Derek Hoiem, Lay-outnet:  Reconstructing the 3D Room Layout from a Single RGB Image, In Proceedings of the IEEE Conference on Computer Vision and Pattern Recognition, pg. 2051–2059, (2018).
%\bibliographystyle{plain}
%\bibliography{ref}
\end{thebibliography}
\end{document}